\pdfoutput=1

\documentclass[11pt]{article}

\usepackage{tcolorbox}  
\tcbuselibrary{breakable}
\tcbset{fontupper=\small}
\usepackage[final]{acl}
\usepackage{enumitem}
\usepackage{times}
\usepackage{latexsym}

\usepackage[T1]{fontenc}

\usepackage[utf8]{inputenc}

\usepackage{microtype}

\usepackage{inconsolata}
\usepackage{titlesec}

\usepackage{graphicx}
\usepackage{amssymb}
\usepackage{multirow}
\usepackage{algorithm}
\usepackage{algorithmic}
\usepackage{amsmath}
\usepackage{booktabs}
\usepackage{xcolor}
\usepackage{subfigure}
\title{A Multi-Expert Structural-Semantic Hybrid Framework for Unveiling Historical Patterns in Temporal Knowledge Graphs}

\author{
    Yimin Deng\textsuperscript{1,2}, 
    Yuxia Wu\textsuperscript{3}, 
    Yejing Wang\textsuperscript{2},
    Guoshuai Zhao\textsuperscript{1}$^{\dagger}$,
    Li Zhu\textsuperscript{1}$^{\dagger}$, 
    Qidong Liu\textsuperscript{1,2}\\
     \textbf{Derong Xu}\textsuperscript{2},
     \textbf{Zichuan Fu}\textsuperscript{2},
     \textbf{Xian Wu\textsuperscript{4}$^{\dagger}$},
     \textbf{Yefeng Zheng}\textsuperscript{5},
     \textbf{Xiangyu Zhao\textsuperscript{2}$^{\dagger}$},
     \textbf{Xueming Qian}\textsuperscript{1}
    \\
    \textsuperscript{1}Xi'an Jiaotong University, \textsuperscript{2}City University of Hong Kong,
    \textsuperscript{3}Singapore Management University\\
    \textsuperscript{4}Tencent Jarvis Lab,
    \textsuperscript{5} Westlake University
    \\
  \small{
  \texttt{
  \href{mailto:dymanne@stu.xjtu.edu.cn}{dymanne@stu.xjtu.edu.cn},
    \href{mailto:guoshuai.zhao@xjtu.edu.cn}{guoshuai.zhao@xjtu.edu.cn},
     \href{mailto:zhuli@xjtu.edu.cn}{zhuli@xjtu.edu.cn}}}\\
     \small{
      \texttt{
    \href{mailto:kevinxwu@tencent.com}{kevinxwu@tencent.com},
    \href{mailto:xianzhao@cityu.edu.hk}{xianzhao@cityu.edu.hk}
  }}
  }

\begin{document}
\maketitle

\begingroup
\renewcommand\thefootnote{\relax}
\footnotetext{$^{\dagger}$  Corresponding authors.}
\endgroup
\begin{abstract}
Temporal knowledge graph reasoning aims to predict future events with knowledge of existing facts and plays a key role in various downstream tasks.~Previous methods focused on either graph structure learning or semantic reasoning, failing to integrate dual reasoning perspectives to handle different prediction scenarios.~Moreover, they lack the capability to capture the inherent differences between historical and non-historical events, which limits their generalization across different temporal contexts.~To this end, we propose a \textbf{M}ulti-\textbf{E}xpert \textbf{S}tructural-\textbf{S}emantic \textbf{H}ybrid (MESH) framework that employs three kinds of expert modules to integrate both structural and semantic information, guiding the reasoning process for different events.~Extensive experiments on three datasets demonstrate the effectiveness of our approach.\footnote{The code and dataset are available at \url{https://github.com/Applied-Machine-Learning-Lab/MESH-TKGR}.} 
\end{abstract}
\section{Introduction}
Incorporating real-world knowledge is essential for enhancing natural language processing capabilities~\cite{xie2023logic,peng2023knowledge,pan2024unifying}.~Directly extracting specific knowledge or facts from various unstructured texts is time-consuming and laborious, hence knowledge graph~(KG) is adopted to store some common facts and reduce retrieval cost.~However, traditional KGs are limited to static fact storage.~To capture the dynamic nature of facts,~temporal knowledge graph~(TKG) was proposed to record facts changing over time.~It can provide certain evidence for many downstream tasks, like situation analysis, political decision making and service recommendation~\cite{mezni2021temporal,saxena2021question,jia2021complex,wu2023resolving,wu2024improving}.
~TKG reasoning aims to predict the missing objects of future events based on existing facts~\cite{leblay2018deriving,garcia2018learning,li2021temporal}, where the formal problem definition is formulated in Section~\ref{problem}.

\begin{figure}[t]
    \centering
    \subfigure[ \vspace{-0.3cm}Methods based on structural information.]{
        \includegraphics[width=\linewidth]{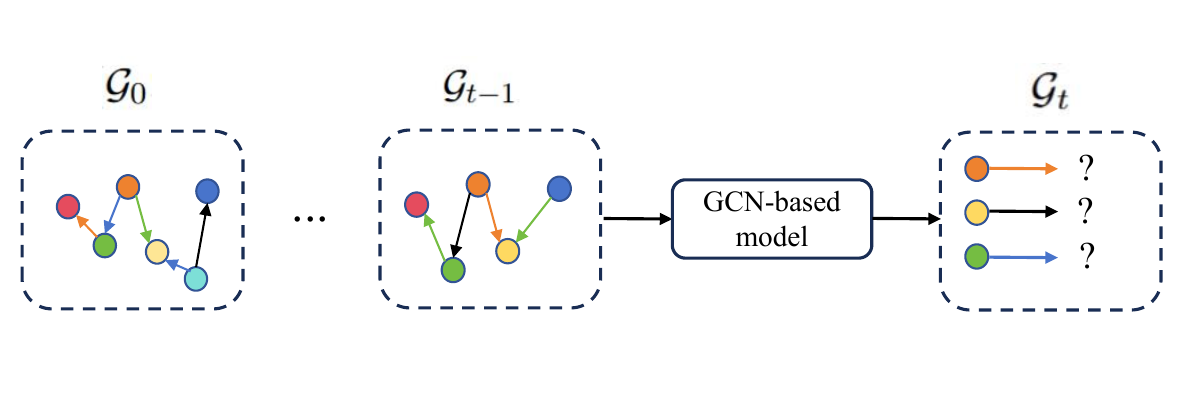}
        \label{setting1}
    }\vspace{-0.3cm}
    \subfigure[ \vspace{-0.3cm}Methods based on semantic information.]{
        \includegraphics[width=\linewidth]{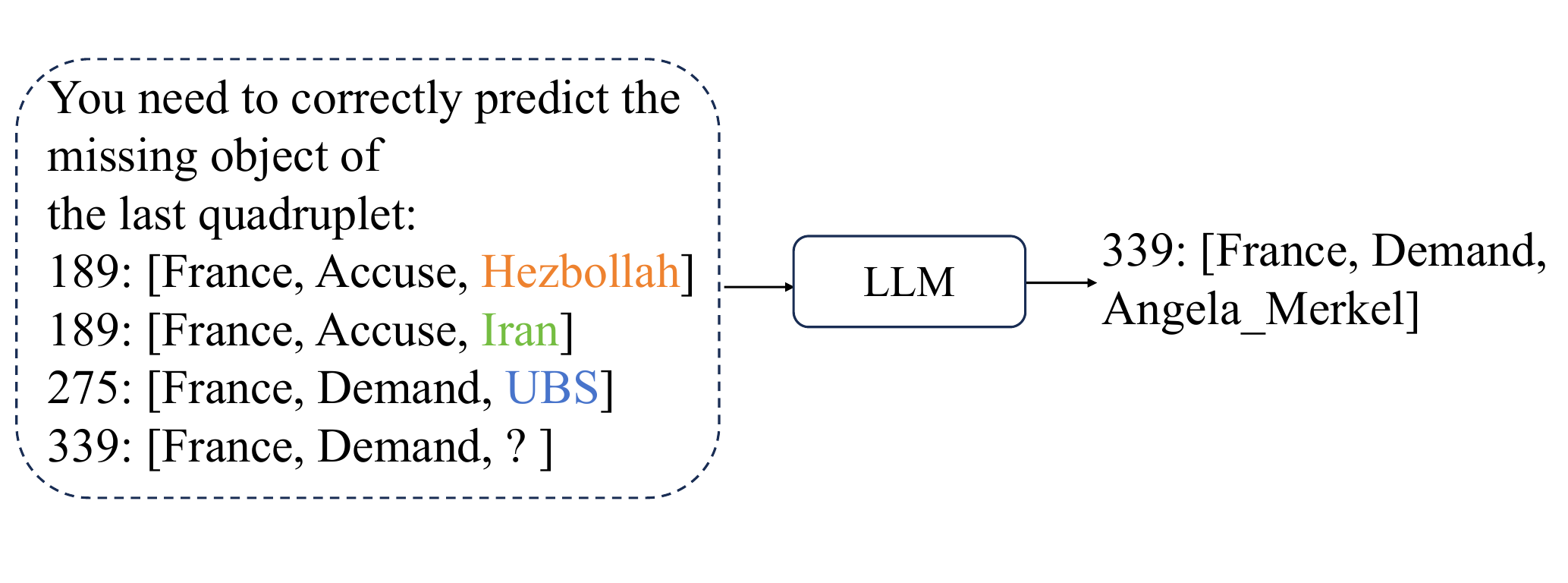}
        \label{setting2}
    }
    \caption{Two research lines of TKG reasoning. One line leverages graph patterns across different timestamps, and the other line utilizes semantic information from event quadruples to capture logical evolution. }
    \label{setting}
    \vspace{-0.5cm}
\end{figure}
According to the type of utilized information, existing methods typically consist of two lines.~One line relies on structural information (Figure~\ref{setting1}), modeling entity interactions through graph neural network~(GCN)~\cite{mo2021survey,cai2023temporal,wang2023survey}.~Previous works in this category have explored the utilization of recurrent networks with neighborhood aggregation mechanisms and recursive graph structures to jointly capture temporal dependencies and concurrent relations in TKGs~\cite{jin2020recurrent,li2021temporal}.~The other line (Figure~\ref{setting2}) focus on semantic reasoning with pre-trained language models~(PLMs)~\cite{xu2023pre}, particularly applying large language models~(LLMs) to generate interpretable reasoning~\cite{chang2024comprehensive}.~These methods~\cite{lee2023temporal,liao2024gentkg,luo2024chain} typically generate predictions through in-context learning with relevant historical facts. Some methods further enhance performance by incorporating retrieval-augmented generation and parameter-efficient tuning, allowing LLMs to better adapt to the TKG reasoning task.~In summary, existing works focus on either structural or semantic information, overlooking the potential benefits of integrating both types of information. However, different types of information can provide complementary insights for reasoning.~The absence of structural information leads to insufficient knowledge of entity interaction patterns, while the lack of semantic information prevents understanding of entities' actions. As exemplified in Figure~\ref{setting2}, France's recurring ``accuse'' or ``demand'' actions could inform predictions about future ``demand'' action.~There is a logical progression from ``accuse'' to ``demand'', but the graph-based methods simply treat them as two different relations, losing this reasoning evidence.

From another viewpoint, events to be predicted can be typically divided into two categories: historical events, which have already taken place in the past, and non-historical events, which have never occurred up to now.~There is an inherent reasoning gap between different events~\cite{xu2023temporal,gastinger2024history}.~For historical events, capturing recurrence patterns is crucial while exploring evolution patterns is essential for non-historical events.~This inspires several works that attempted to handle historical and non-historical events differently.~TiRGN~\cite{li2022tirgn} models both the periodic patterns~(which often appear in recurring historical events) and sequential evolution patterns~(which characterize non-historical events), while CENET~\cite{xu2023temporal} employs a binary classifier to separate the two types of events and focus predictions on relevant candidate entities.
~These methods often overlook that different types of information have their own advantages when handling different types of events.
To address the aforementioned limitations, we propose a \textbf{M}ulti-\textbf{E}xpert \textbf{S}tructural-\textbf{S}emantic \textbf{H}ybrid (MESH) framework that effectively integrates structural and semantic information to model historical patterns for temporal knowledge graph reasoning.
This model consists of a feature encoder, two kinds of event-aware expert modules and a prediction expert module.~The underlying feature encoder outputs structural information from GCN and semantic information from LLM.~Then we employ two kinds of event-aware expert modules to learn information weight allocation patterns for historical/non-historical events.~There is a challenge in distinguishing event types.
~To address this, we design a prediction expert module which assigns different weights to each event-aware expert module, thereby implicitly distinguishing different types of events. This unified architecture enables adaptive information fusion without requiring explicit event type labels, offering both flexibility and efficiency.
To summarize, the contributions of our work are as follows:
\begin{itemize}[leftmargin=*,nolistsep]
\item We discover and verify the complementary advantages of structural and semantic information when applied to different types of events.
\item We propose a novel non-generative approach to leveraging LLMs for TKG reasoning, in combination with graph-based models. 
\item 
We employ two kinds of event-aware expert modules that adapt to different information preferences between historical and non-historical events, with a prediction expert for automatic weight allocation between experts.
\item We conduct extensive experiments on three public benchmarks and the results demonstrate the effectiveness of our proposed method.
\end{itemize}

{\titleformat{\subsubsection}[runin]
    {\normalfont\bfseries}
    {\thesubsubsection}
    {1em}
    {}
\section{Related Work}
\subsubsection*{GCN-Based TKG Reasoning Models.}
GCNs have shown strong ability to model structural information for graphs, leading to a series of GCN-based methods for temporal knowledge reasoning.~RE-Net~\cite{jin2020recurrent} utilizes a recurrent event encoder and neighborhood aggregator to capture temporal dependencies and concurrent relations. REGCN~\cite{li2021temporal} recurrently fit entity and relation features in the order of timestamps.
TiRGN~\cite{li2022tirgn} explicitly integrates time embeddings to graph embeddings, which facilitates learning across long temporal periods.
~GCN-based models treat entities as nodes and integrate representations of entities and relations to predict the next event. While effective in capturing structural patterns, these methods often overlook semantic information in the reasoning process. 

\subsubsection*{LLM-Based TKG Reasoning Models.}
Some methods formulate TKG reasoning as masked language modeling~(MLM) or next-token generation tasks, using language models as the backbone.~ChapTER~\cite{peng2024deja} uses a pre-trained language model as encoder and employs a prefix-based tuning method to obtain good representations.~PPT~\cite{xu2023pre} performs masked token prediction with fine-tuned BERT.~Recently, LLMs have demonstrated powerful capabilities in summarization and reasoning.~ICL~\cite{lee2023temporal} leverages LLMs with in-context learning by carefully selecting historical facts as context and decodes the outputs to rank predicted entities.~GenTKG~\cite{liao2024gentkg} proposes a retrieval-augmented generation approach with temporal logic rules, while applying parameter-efficient tuning to adapt LLMs for TKG reasoning.~CoH~\cite{luo2024chain} captures key historical interaction from input text and also applies parameter-efficient tuning.~Although these LLM-based methods provide context-based interpretable reasoning, they often struggle to capture the complex structural patterns in TKGs.
}
\section{Methods}

In this section, we first introduce the problem definition and notations of the temporal knowledge graph reasoning. Then we present the overall architecture of our proposed approach, followed by a detailed description of each model component. 
\subsection{Problem Formulation}\label{problem}

A TKG $\mathcal{G}$ consists of a set of static graphs $\mathcal{G}_t$, where each static graph contains all the facts at timestamp $t$.~Formally, a TKG can be represented as $\mathcal{G}$ = $\{\mathcal{E, R, T, F}\}$, where  $\mathcal{E}$, $\mathcal{R}$, and $\mathcal{T}$ denote the sets of entities, relations, and timestamps, respectively.
~$\mathcal{F}$ denotes the set of facts, each formulated as a quadruple $(s, r, o, t)$, where $s,o \in \mathcal{E}$, $r \in \mathcal{R}$, and $t \in \mathcal{T}$, $s$/$o$ represents the subject/object and $r$ represents the relation between $s$ and $o$ at $t$. 
To model the dynamic nature of real-world knowledge, the TKG reasoning task aims to predict the missing entity in query
$(s, r, \_, t)$ 
with existing facts occur before time $t$. 
Additionally, an event $(s,r,o,t)$, if there exists a previous occurrence $(s,r,o,k)$ where $k<t$, is denoted by historical event; otherwise, is denoted by non-historical event. 

\subsection{Overall Framework}
\begin{figure}[t]
    \centering
    \resizebox{1\linewidth}{!}{
    \includegraphics[height=0.6\textwidth]{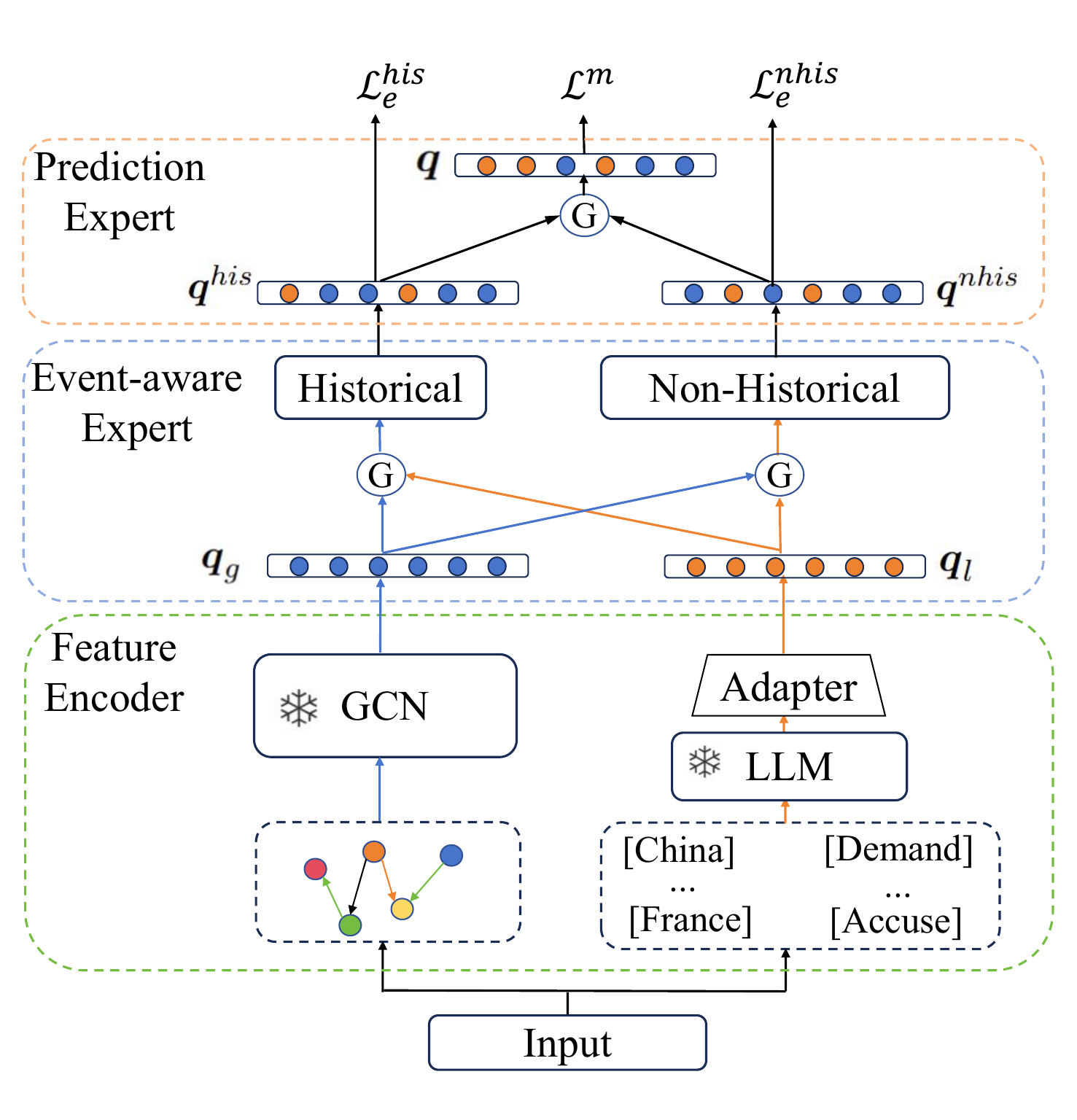}}
    \caption{The overall architecture of MESH. 
    }
    \label{model}
    \vspace{-3mm}
\end{figure}
In this section, we briefly introduce the overall architecture of MESH. As shown in Figure~\ref{model}, it follows a three-layer architecture, consisting of an underlying feature encoder, two kinds of event-aware expert modules in the middle layer, and a prediction expert at the top.~Specifically, the feature encoder contains two components: a GCN-based structural encoder taking sub-graph structures as input and a LLM-based semantic encoder taking prompts with entity/relation names as input.~These two components capture TKG information from complementary perspectives and generate query representations $\boldsymbol{q}_g$ and $\boldsymbol{q}_s$, respectively.~To adaptively handle feature fusion at different layers, we employ query-motivated gates that take $\boldsymbol{q}_g$ as input.~The two kinds of specialized event-aware expert modules control the fusion patterns of features for historical or non-historical events, producing representations $\boldsymbol{q}_{his}$/$\boldsymbol{q}_{nhis}$, and the prediction expert learns the representation $\boldsymbol{q}$ for final prediction.



\subsection{Feature Encoder}
 
 High-quality encoders are essential for integrating and analyzing dynamic knowledge~\cite{zhao2018deep,zhao2018recommendations,zhang2023autostl,wang2023single}.~Since different types of encoders can capture complementary perspectives of TKGs, we employ two independent encoders: a GCN-based encoder for extracting information from graph structures and an LLM-based encoder for modeling semantic information. In this section, we will introduce these two encoders in detail.
 \subsubsection{Structural Encoder} 

 We employ a structural encoder that learns expressive representations of entity interactions over time. The structural encoder aggregates relational information from the graph topology, enabling the model to incorporate both structural dependencies and temporal dynamics~\cite{xu2024multi}. Formally, given a temporal knowledge graph $\mathcal{G}$ the structural encoder $G$ generates structural embeddings as:
 \begin{equation}
 \label{eq1}
 \boldsymbol{H}_{g}, \boldsymbol{R}_{g} = G(\mathcal{G})
\end{equation}
where $\boldsymbol{H}_{g}\in \mathbb{R}^{|\mathcal{E}|\times d}$ and $\boldsymbol{R}_{g}\in \mathbb{R}^{|\mathcal{R}|\times d}$ denote the structural feature of entities and relations, respectively.
In this paper, we adopt RE-GCN \cite{li2021temporal} as the structural encoder, but our method is not restricted to any specific structural encoder.

\subsubsection{Semantic Encoder}

Entities and relations in TKGs usually contain rich semantic information.~For example, the entity ``Javier Solan'' is associated with semantic attributes as a Spanish politician, providing valuable contextual knowledge crucial for reasoning tasks. Leveraging this semantic information is essential for enhancing TKG reasoning and improving model performance. Existing methods often leverage the reasoning and generative capabilities of LLMs to directly generate the answer for TKG reasoning tasks~\cite{lee2023temporal,liao2024gentkg}. However, they typically suffer from high inference latency. To mitigate this issue, recent works have focused on leveraging the representational capacity of LLMs to reduce inference costs~\cite{wang2023multi,wang2023plate,liu2024llm}. Motivated by this, we adopt an LLM-based approach to encode entities and relations efficiently.

Specifically, we construct the following prompts: 
\begin{tcolorbox}[
        colframe=brown,
        width=1\linewidth,
        arc=1mm, 
        auto outer arc,
        title={\small Entity Encoding Template},
        breakable,]
        In the context of \underline{<DATA DOMAIN>}, please provide \underline{<DATA TYPE>} background about \underline{<ENTITY>}. 
\end{tcolorbox}
\begin{tcolorbox}[colframe=brown,
        width=1\linewidth,
        arc=1mm, 
        auto outer arc,
        title={\small Relation Encoding Template},
        breakable,]
       In the context of \underline{<DATA DOMAIN>}, what are the \underline{<DATA TYPE>} perspectives through which we can understand the \underline{<RELATION>}?
\end{tcolorbox}
We fill the underlined places with dataset-specific characteristics and particular entities or relations, then feed these prompts to LLMs such as LLaMA \cite{touvron2023llama} to obtain semantic embeddings.
For example, for ICEWS14 dataset, \underline{<DATA DOMAIN>} and \underline{<DATA TYPE>} can be political and historical wordings. \underline{<ENTITY>} can be `France' and \underline{<RELATION>} can be `Abuse'. Finally, we extract the hidden states from the last transformer layers of the LLM to obtain semantic representations of entities and relations, denoted as $\boldsymbol{H}_{LLM}\in \mathbb{R}^{|\mathcal{E}|\times d_{LLM}}$ and $\boldsymbol{R}_{LLM}\in \mathbb{R}^{|\mathcal{R}|\times d_{LLM}}$.~However, the original LLM representations are trained for general language tasks~\cite{touvron2023llama} and typically have significantly larger dimensions than structural embeddings($d_{LLM}>>d$). To adapt $\boldsymbol{H}_{LLM}, \boldsymbol{R}_{LLM}$ to our TKG reasoning tasks, we employ adapter modules  $f_H,f_R$ that compress these representations to a lower-dimensional space, typically implemented as multi-layer perceptrons (MLPs)~\cite{chen2024enhance,liu2024moe}:
\begin{align}
\label{eq2}
\boldsymbol{H}_{l}=f_H(\boldsymbol{H}_{LLM}),\boldsymbol{R}_{l}=f_R(\boldsymbol{R}_{LLM})
\end{align}
where $\boldsymbol{H}_{l}\in \mathbb{H}^{|\mathcal{E}|\times d},\boldsymbol{R}_{l}\in \mathbb{R}^{|\mathcal{E}|\times d}$.

\subsubsection{Query Representation}
From the underlying feature encoder, we obtain entity embeddings $\boldsymbol{h}_g\in \mathbb{R}^d$, $\boldsymbol{h}_l\in \mathbb{R}^d$ and relation embeddings $\boldsymbol{r}_g\in \mathbb{R}^d$, $\boldsymbol{r}_l\in \mathbb{R}^d$ by taking the corresponding rows from $\boldsymbol{H}_{g},\boldsymbol{H}_{l},\boldsymbol{R}_{g},\boldsymbol{R}_{l}$.~We subsequently detail decoders to generate query representations for the query $(s,r,\_,t)$. 

Since the convolutional score function has shown its effectiveness as decoder in previous work~\cite{vashishthcomposition,li2021temporal}, we employ ConvTransE as decoder:
\begin{gather}
    \boldsymbol{q}_g=D_{g}(\boldsymbol{h}_g \oplus \boldsymbol{r}_g)\label{eq3}
\\
    \boldsymbol{q}_l=D_{l}(\boldsymbol{h}_l \oplus \boldsymbol{r}_l)\label{eq4}
\end{gather}
where $D_g$ and $D_l$ denote the decoders for structural and semantic features, respectively, $\boldsymbol{q}_g\in \mathbb{R}^{d}$ and $\boldsymbol{q}_l\in \mathbb{R}^{d}$ denote the query representations from structural and semantic perspectives.

\subsection{Event-Aware Experts}
We suggest that different type of events may require different kind of information for reasoning. For example, historical events often involve complex context which are better captured by LLMs, as demonstrated by our analysis in Section~\ref{his/nhis}. Existing methods~\cite{li2021temporal,liao2024gentkg} overlook this diversity and thus fail to achieve optimal performance on different types of events consistently. This observation motivates us to design a mechanism that can adaptively handle different types of events. Consequently, we propose event-aware experts in this section to adaptively integrate structural and semantic information from previous steps. As shown in Section~\ref{ablation}, it effectively enhances the model's reasoning capability.

Specifically, we divide events into two categories: historical and non-historical~\cite{xu2023temporal}. We set $M$ experts for historical events and $N$ experts for non-historical events~\cite{zhang2024m3oe}.~We employ $\boldsymbol{q}_g$ as the input to the query-motivated gate, since it captures the evolving structural patterns of each sub-graph over time and records dynamic structural information, thereby can better distinguish event types. The operation of $i^{th}(i\leq M+N)$ expert is:
\begin{gather}
\alpha_i = \sigma(\boldsymbol{q}_g\mathbf{W}_i + b_i)\label{eq:ev-aware-gate}\\
\boldsymbol{q}_i = \alpha_i\cdot \boldsymbol{q}_g + (1-\alpha_i) \cdot \boldsymbol{q}_s\label{eq:ev-aware-experts}
\end{gather}
where 
$\mathbf{W}^i\in \mathbb{R}^{d\times1}$ and $b^i$ denote the weight matrix and bias term of the gating function to generate the weight $\alpha_i$ that indicating the dependency of this expert on structural information.~$\boldsymbol{q}_i\in \mathbb{R}^{d}$ represents the query representation from the $i$-th expert module, which combines comprehensive views for further prediction. We regard the first $M$ experts as historical experts and the following $N$ as non-historical experts, which are adept in handling historical and non-historical events, respectively.

\subsection{Prediction Expert}

While we have studied how to leverage different information based on event types, distinguishing the types of events to be predicted remains challenging since the types of future events are unknown. Previous methods~\cite{xu2023temporal} typically employ binary classification to determine event types, making the prediction performance susceptible to classification errors. To address this limitation, we design a prediction expert that adaptively integrates information from different kinds of experts without explicit type classification.

Based on query representations produced by multiple event-aware experts, we finally construct a prediction expert to mix all information and predict future events. Similar to Equation~\eqref{eq:ev-aware-gate}, we first adaptively allocate weights to experts with the initial query representation $\boldsymbol{q}_g$:
\begin{gather}  
    \boldsymbol{\alpha}=\sigma(\boldsymbol{q}_g\boldsymbol{W} + \boldsymbol{b})\label{eq7}\\
    \boldsymbol{q} =\boldsymbol{\alpha}  \cdot [\boldsymbol{q}_1,...,\boldsymbol{q}_{M+N}]^T=\sum_{1\leq i\leq M+N} \alpha_i\boldsymbol{q}_i\label{eq:predict-dot}
\end{gather}
where $\mathbf{W}\in \mathbb{R}^{d\times {(M+N)}}$ and $\mathbf{b} \in \mathbb{R}^{(M+N)}$ denote the weight matrix and bias term of the gating function, resulting in $\boldsymbol{\alpha}\in\mathbb{R}^{(M+N)}$. Equation~\eqref{eq:predict-dot} combines expert information with the dynamic weights. 


The final prediction $\boldsymbol{p}_{s,r,t}$ for query $(s,r,\_,t)$ is made with the matrix product between $\boldsymbol{q}$ and $\boldsymbol{H}_g$:
\begin{gather}
    \boldsymbol{p}_{s,r,t}=\sigma(\boldsymbol{q}\cdot \boldsymbol{H}_g) \label{eq:finalP}
\end{gather}
where $\boldsymbol{p}_{s,r,t} \in \mathbb{R}^{|\mathcal{E}|}$ indicates the probability of the missing object is corresponding candidate within the candidate set $\mathcal{E}$.

\begin{table*}[t]
\centering
\small
\resizebox{0.9\textwidth}{!}{
\begin{tabular}{c c c c c c c c c c}
\hline

\hline
             Dataset&$|\mathcal{E}|$ &$|\mathcal{R}|$&Train&Valid&Test&$|\mathcal{F}_{his}|$ &$Rate_{his}$& $\Delta t$        \\ \hline
ICEWS14   &  7,128 & 230 &74,845&8,514&7,371&3,064&41.6\%&24h \\ 
ICEWS18 & 23,033 &256&373,018& 45,995& 4,9545&20,865&42.1\%&24h \\
ICEWS05-15 &  10,778 &24 &368,868 &46,302&46,159&24,915&54.0\%&24h \\
\hline

\hline
\end{tabular}}
\caption{Statistics of datasets. $|\mathcal{F}_{his}|$ and $Rate_{his}$ denote the number and percentage of historical events in test set, respectively, and $\Delta t$ denotes the time granularity of each dataset.}
\label{tab:dataset}
\end{table*}

\subsection{Optimization}
In this section, we introduce the loss function for the model optimization.~There are two training objectives: 1) Event-aware experts should specialize in corresponding event types. 2) The overall prediction is accurate for TKG reasoning tasks. 

We can obtain the partial information of historical and non-historical experts as: 
\begin{gather}
    \boldsymbol{q}^{his} = \sum_{1\leq i\leq M} \alpha_i\boldsymbol{q}_i\label{eq10}\\ 
    \boldsymbol{q}^{nhis} = \sum_{M+1 \leq i\leq M+N} \alpha_i\boldsymbol{q}_i\label{eq11}
\end{gather}
Then, the event-aware predictions can be obtained  similar to Equation~\eqref{eq:finalP}:
\begin{gather}
    \boldsymbol{p}_{s,r,t}^{his}=\sigma(\boldsymbol{q}^{his}\cdot \boldsymbol{H}_g)\label{eq12}\\
    \boldsymbol{p}_{s,r,t}^{nhis}=\sigma(\boldsymbol{q}^{nhis}\cdot \boldsymbol{H}_g)\label{eq13}
\end{gather}
By definition, historical facts refer to events that have occurred before $t$, 
while non-historical facts have no prior occurrences at time $t$. We can calculate the frequency of each fact's occurrence before $t$ and construct the historical indicator:
\begin{gather}
  F^{s,r}_t(o)=\sum_{k<t}\lvert\{(s,r,o,k)|(s,r,o,k)\in \mathcal{G}_k\}\rvert \\
  I^{s,r}_t(o)=\begin{cases}
      1\text{  if } F^{s,r}_t(o)>0\\
      0\text{  if } F^{s,r}_t(o)=0
  \end{cases}\label{eq15}
\end{gather}
where $\mathcal{G}_k$ denotes the subgraph at time $k$. The set $\{(s,r,o,k)|(s,r,o,k)\in \mathcal{G}_k\}$ indicates all events happened at timestamp $k$ that contains the object $o$. $F^{s,r}_t(o)$ denote the summation of the count over these set prior to the current timestamp $t$, indicating the frequency of the event. The historical indicator $I^{s,r}_t(o)$ can finally distinguish event types if the frequency is non-zero.

To enable event-aware expert modules to specialize in different event types, we compute the expert loss based on corresponding predictions:
\begin{gather}
  \mathcal{L}^{his}_e=-\sum_{(s,r,o,t)\in \mathcal{G}}\boldsymbol{y}_{s,r,t} \boldsymbol{p}_{s,r,t}^{his}I^{s,r}_t(o)\label{eq16} \\
  \mathcal{L}^{nhis}_e=-\sum_{(s,r,o,t)}\boldsymbol{y}_{s,r,t} \boldsymbol{p}_{s,r,t}^{nhis}(1-I^{s,r}_t(o))\label{eq17}
\end{gather}
where $\boldsymbol{y}_t^{s,r}\in \mathbb{R}^{|\mathcal{E}|}$ is the one-hot ground truth vector of for entity prediction. 

Besides, the prediction with all experts should be accurate, resulting in the major loss:
\begin{equation}
    \mathcal{L}^m=-\sum_{(s,r,o,t)\in \mathcal{G}}\boldsymbol{y}_{s,r,t} \boldsymbol{p}_{s,r,t}\label{eq18}
\end{equation}
The total loss function is formulated as a weighted sum of the major loss and the auxiliary expert loss:
\begin{equation}
    \mathcal{L}=\mathcal{L}^m +\omega (\mathcal{L}^{his}_e+\mathcal{L}^{nhis}_e)\label{eq19}
\end{equation}
where $\omega$ is the balancing weight. We detail the training procedure in \textbf{Appendix~\ref{app:Opt}}.

\section{Experiments}
\subsection{Experimental Setup}
\subsubsection{Datasets}
We conduct experiments on three public benchmarks ICEWS14~\cite{garcia2018learning}, ICEWS18~\cite{jin2020recurrent} and ICEWS05-15~\cite{garcia2018learning}.
 ~Table~\ref{tab:dataset} shows the statistics of these datasets. All these three datasets show a balanced distribution between historical and non-historical facts in their test sets, with historical event ratios ranging from 41.6\%~(ICEWS14) to 54.0\%~(ICEWS05-15), making them suitable for evaluating the prediction of different event types.
\subsubsection{Evaluation Metrics}
Following previous works~\cite{jin2020recurrent,li2021temporal}, we employ the Mean Reciprocal Rank~(MRR) and Hits@k~(H@k) as the evaluation metrics.~MRR calculates the average of the reciprocal ranks of the first relevant entity retrieved by the model, while H@k calculates the proportion of the correct entity ranked in the top k.~The MRR metric is not available for LLM-based methods as they directly generate entities rather than ranking candidates.~Detailed definitions of these metrics are shown in \textbf{Appendix~\ref{metric}}.

\subsubsection{Baselines}
To conduct a comprehensive comparison, we select nine up-to-date TKG reasoning methods, including five graph-based methods and four LLM-based methods.~For graph-based models, we choose RE-Net~\cite{jin2020recurrent}, REGCN~\cite{li2021temporal},~CENET~\cite{xu2023temporal}.
For LLM-based models, we choose ICL~\cite{lee2023temporal}, CoT~\cite{luo2024chain} and GenTKG~\cite{liao2024gentkg}.
We also compare with two straightforward baselines introduced in \textbf{Appendix~\ref{naive}}, namely ``Naive'' and ``LLM-MLP''.


\begin{table*}[!t]
 \centering
 \setlength{\tabcolsep}{1.4mm}
\resizebox{0.9\linewidth}{!}{
 \begin{tabular}{lccccccccc}\toprule
    \multirow{2}{*}{\textbf{Model}} & \multicolumn{3}{c}{\textbf{ICEWS14}} & \multicolumn{3}{c}{\textbf{ICEWS18}} & \multicolumn{3}{c}{\textbf{ICEWS05-15}} 
    \\\cmidrule(lr){2-4}\cmidrule(lr){5-7}\cmidrule(lr){8-10}
             & MRR & H@3 & H@10  & MRR & H@3 & H@10  & MRR & H@3 & H@10  \\\midrule \specialrule{0em}{1.5pt}{1.5pt}
    \textbf{\textit{Graph-Based Methods}} \\
    Naive & - & 38.00 & 44.73 & - & 4.04 & 6.29 & - & 39.66 & 49.68\\
     RE-Net~\cite{jin2020recurrent} & 40.89&45.31 &59.23 &29.92&32.44&49.51&43.55&48.86&64.78  \\
    RE-GCN~\cite{li2021temporal} & 41.89 & 46.26 & 61.4 & 32.41 & 36.74 & 52.27&48.42 & 53.97 & 68.47  \\
    TiRGN~\cite{li2022tirgn} & \underline{44.06} & \underline{49.02} &\underline{63.84}  &\underline{33.42}  & \underline{37.87} & \underline{54.11}& \textbf{49.61}&  \textbf{55.49}&\textbf{69.91}   \\
    CENET~\cite{xu2023temporal}&39.92  &43.56  &58.2  & 27.98 &31.67  &47.02 &42.0 &46.99  & 62.3  \\
    \midrule
    \textbf{\textit{LLM-Based Methods}} \\
    LLM-MLP & 39.77&43.62&58.69  & 29.25&32.88&47.65  & 39.85& 43.5 &58.84   \\
    LLM-ICL~\cite{lee2023temporal} & -&38.9&45.6  & - &29.7& 36.6 &- & 49.1 &  56.6 \\
    
    GenTKG~\cite{liao2024gentkg} &- &48.3& 53.6&-& 37.25&49.9&-&52.5&68.7 \\
    CoH~\cite{luo2024chain} & - & 46.64 & 59.11 &-  &36.22& 52.61 & - &53.14  & \underline{68.87}  \\
    \midrule
    MESH & \textbf{44.36} & \textbf{49.81} & \textbf{64.21} & \textbf{33.96} & \textbf{38.37} & \textbf{54.12} & \underline{48.66} & \underline{54.26}&68.57
    \\\bottomrule
 \end{tabular}}
 \caption{TKG reasoning performance (with time-aware metrics) on ICEWS14, ICEWS18, ICEWS05-15. The best results are in \textbf{bold} and the second best results are \underline{underlined}. Results are averaged over three random runs (p $<$ 0.05 under t-test).
 }
 \label{tab:main results transductive}
 \vspace{-2mm}
\end{table*}

\subsubsection{Implementation Details}
For the structural encoder $G$, we employ an efficient graph-based model RE-GCN~\cite{li2021temporal}. The hidden dimension $d$ is 100. The dropout of each GCN layer is set to 0.2. To maintain the stability of structural features, the structural encoder is trained for 500 epochs and then frozen. For the semantic encoder, we utilize LLaMA-2-7B ($d_{LLM}=4096$) coupled with a two-layer MLP as the adapter.~The decoder $D_g$ and $D_l$ are ConvTransE with 50 channels and kernel size of 3. For event-aware experts, $M=N=1$. 
All experiments are conducted on an NVIDIA A100 GPU, with the learning rate set to 0.001. Our results are averaged over three random runs.

\subsection{Main Results}
The experimental results for entity prediction are presented in Table \ref{tab:main results transductive}. Based on these results, we observe the following findings:
\begin{itemize}[leftmargin=*, nolistsep]

\item MESH achieves state-of-the-art performance on ICEWS14 and ICEWS18, surpassing all graph-based and LLM-based baselines.~It maintains competitive performance on ICEWS05-15, only second to TiRGN~\cite{li2022tirgn}. Moreover, our proposed model can be integrated with any structural or semantic encoder, improving methods like TiRGN as shown in Table~\ref{tab:compatibility study}. 

\item MESH significantly outperforms our structural encoder RE-GCN on all results. After incorporating semantic information, our model outperforms RE-GCN with improvements of 2.47\%/1.55\% in MRR, 3.55\%/1.63\% in H@3, and 2.81\%/1.85\% in H@10 on ICEWS14/ICEWS18,~respectively.~These results demonstrate that the strong understanding capability of LLMs can effectively enhance the model's power of prediction.

\item Compared to LLM-based methods, our approach demonstrates superior performance across all available metrics.~It demonstrates the importance of structural information in TKG reasoning. Notably, LLM-based methods show consistently lower scores on H@10 compared to graph-based approaches, revealing their inherent limitations in maintaining comprehensive historical knowledge. This observation aligns with the catastrophic forgetting during continual learning of LLM, indicating their difficulty in effectively leveraging complete historical patterns during reasoning. 
\item Our method addresses this limitation of LLM-based model by integrating structural information with global information of graph. Moreover, our model significantly reduces the inference cost. While these LLM-based methods usually require over 12 hours for inference on ICEWS14, our approach completes the same task within minutes.

\end{itemize}


\subsection{Compatibility Study}
To validate the compatibility of MESH, we conduct experiments with different structural and semantic encoders, as shown in Table \ref{tab:compatibility study}. Our default configuration employs RE-GCN as the structural encoder $G$ in Equation~\eqref{eq1} and LLaMA2-7B as the semantic encoder in Equation~\eqref{eq2}.~For structural encoders, TiRGN shows superior performance over RE-GCN across all metrics. For semantic encoders, Stella-en-1.5B-v5~\cite{zhang2024jasper} slightly outperforms LLaMA2-7B. When integrating these alternative encoders into our framework, we observe consistent improvements. Specifically, replacing RE-GCN with TiRGN leads to better performance, achieving an MRR of 44.97\%, H@3 of 50.78\%, and H@10 of 65.54\%.~Incorporating Stella also brings performance improvements. Since structural encoders (e.g., RE-GCN, TiRGN) generally outperform semantic encoders (e.g., LLaMA2, Stella), the improvements are less significant compared to those obtained from better structural encoders.~Overall, our method achieves consistent performance gains by integrating both structural and semantic encoders, compared to using a single encoder alone.~These experimental results strongly support our claim that our model is not limited to specific structural or semantic encoder, allowing MESH to effectively integrate various advanced encoder modules. 
\begin{table}[t]
\centering
\resizebox{0.45\textwidth}{!}{
\begin{tabular}{cccc}
\hline
               & MRR   & H@3   & H@10    \\ \hline
RE-GCN& 41.89 & 46.26 & 61.4\\
TiRGN  & 44.06 & 49.02 &63.84  \\\hline
 LLama-2-7B&39.77 & 43.62 & 58.69 \\ 
  Stella-en-1.5B-v5&40.39  & 44.60 & 58.96 \\ \hline
   MESH   & 44.36 & 49.81 &64.21  \\ 
   MESH w/TiRGN   & \underline{44.97} & \textbf{50.78} &\underline{ 65.54}  \\
   MESH w/Stella  & 44.53 & 49.23 & 63.53 \\
   MESH w/TiRGN, Stella   & \textbf{45.05} & 50.56 &\textbf{ 65.77}  \\\hline
\end{tabular}}
\vspace{-0.2cm}
\caption{Compatibility study on ICEWS14.}
\label{tab:compatibility study}
\vspace{-0.3cm}
\end{table}
\begin{figure}[t]
    \centering
    \includegraphics[width=\linewidth]{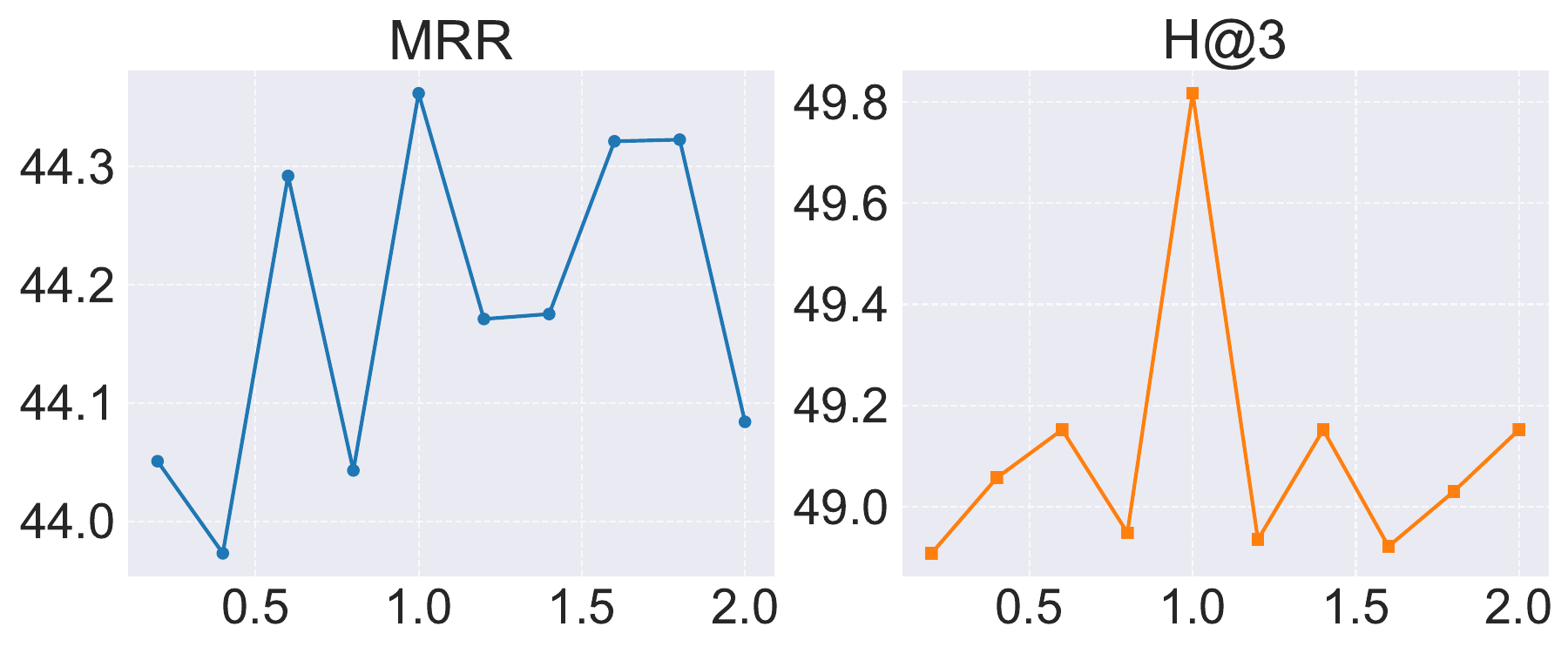}
    \caption{Sensitivity analysis results of $\omega$ on ICEWS14. 
    }
    \label{fig:weight_impact}
    \vspace{-2mm}
\end{figure}
\begin{table}[t]
\centering
\resizebox{0.45\textwidth}{!}{
\begin{tabular}{cccc}
\hline
               & MRR   & H@3   & H@10    \\ \hline
w/o Semantic Info& 41.89 & 46.26 & 61.4 \\
w/o Structural Info     &39.77 & 43.62 & 58.69 \\ 
w/o  Event-aware  & 43.96 & 48.92 &64.15  \\
w/o  Prediction Expert  &43.44  & 48.28 & 62.77 \\ \hline
MESH   & \textbf{44.36} & \textbf{49.81} &\textbf{ 64.21}  \\ \hline
\end{tabular}}
\caption{Ablation study on ICEWS14.}
\label{tab:ablation study}
\vspace{-0.3cm}
\end{table}

\begin{table}[t]
 \centering
 \setlength{\tabcolsep}{1.9mm}
 \resizebox{\linewidth}{!}{
 \begin{tabular}{ccccccc}\toprule
    \multirow{2}{*}{\textbf{Model}} & \multicolumn{3}{c}{\textbf{Historical}} & \multicolumn{3}{c}{\textbf{Non-Historical}} 
    \\\cmidrule(lr){2-4}\cmidrule(lr){5-7}
             & MRR & H@3 & H@10  & MRR & H@3 & H@10  
             \\\midrule \specialrule{0em}{1.5pt}{1.5pt}
   
   
    CENET	&68.32	&72.68&	86.77&15.02	&17.22	&30.3\\
   RE-GCN &66.75 &73.65& 87.10& 24.22 &26.79& 43.25 
   \\\hline
     LLM-ICL &-&77.2&82.7&-	&0.1	&0.1\\
    GenTKG &- &\textbf{82.1}& 86.8&-& 16.9&22.8
    \\
    
    \midrule
    MESH & \textbf{72.81} & 80.80 & \textbf{91.81} & \textbf{24.52} & \textbf{27.13} & \textbf{44.06} 
    \\\bottomrule
 \end{tabular}}
 \caption{Different event types on ICEWS14. }
 \label{tab:historical results}
\end{table}

\begin{table}[t]
\centering
\resizebox{0.35\textwidth}{!}{
\begin{tabular}{cccc}\toprule
$\alpha_1$& Historical & Non-Historical \\\hline
Mean & 0.5341 & 0.4589 \\
Std & 0.079 & 0.094 \\
p-value & \multicolumn{2}{c}{$<$ 0.001} \\
\bottomrule
\end{tabular}}%
\caption{T-test for $\alpha$ on ICEWS14}
\vspace{-3mm}

\label{tab:case}
\end{table}
\subsection{Ablation Study}
\label{ablation}
Table \ref{tab:ablation study} shows the ablation studies of our proposed model.
First, we remove the semantic or structural information obtained from Equation~\eqref{eq1}/\eqref{eq2}, denoted as \textbf{w/o Semantic Info} or \textbf{w/o Structural Info}. It leads to a 2.47\%/4.59\% decrease in MRR, indicating that both types of information are complementary and crucial for accurate predictions. Next, we drop the specialization of event-aware experts as \textbf{w/o Event-aware}, specifically by removing the auxiliary expert loss in Equation~\eqref{eq19}. 
Finally, we omit the prediction expert, i.e., replace $\boldsymbol{q}$ in Equation~\eqref{eq:finalP} with the average of $\{\boldsymbol{q}_i\}$, denoted by \textbf{w/o Prediction Expert}. It leads to decreases of 0.92\% in MRR, 1.53\% in H@3, and 1.54\% in H@10,~indicating the importance of adaptively integrating multiple event-aware experts. \textbf{Appendix~\ref{app:ablation}} provides the further studies on gating inputs.

\vspace{-2mm}
\subsection{Sensitivity Analysis}
To explore the sensitivity of MESH to the expert loss weight $\omega$ in Equation~\eqref{eq19}, we conduct experiments by varying the value of $\omega$ from 0.2 to 2.0.~As shown in Figure~\ref{fig:weight_impact}, we evaluate the model performance on ICEWS14 using MRR and H@3. The MRR varies between 43.97\% and 44.36\%, while H@3 varies between 48.91\% and 49.81\%.~The results show that our model maintains stable performance across different values of $\omega$. As $\omega$ increases, the model performance first improves and then declines, achieving the best results when $\omega=1$.~This trend indicates that the model performs best when the expert loss and prediction loss are weighted equally, showing that maintaining a balanced scale between these two loss components is better. 


\subsection{In-depth Analysis on Event Types}\label{his/nhis}
In this section, we conduct three experiments to validate our claim that different event types require distinct types of information, and to explore the optimal number of experts.

\vspace{+0.1cm}\noindent\textbf{Performance on Different Events.}
As shown in Table \ref{tab:historical results}, we observe distinct performance between graph-based methods and LLM-based methods on different types of events.~Graph-based methods like RE-GCN demonstrate strong capability in capturing evolution patterns through their structural modeling, while LLM-based models (e.g., GenTKG) excel particularly at modeling historical events due to their powerful representation learning but show limited generalization to non-historical scenarios.
Our proposed method achieves consistent improvements in both scenarios, suggesting its effectiveness in learning specific reasoning patterns for different types of events.~This balanced performance can be attributed to our model's ability to leverage both structural patterns and semantic information effectively, bridging the gap between historical/non-historical events.

\vspace{+0.1cm}\noindent\textbf{Statistic Analysis of Prediction Expert.}
In this part, we present a statistical analysis to demonstrate the ability of the prediction expert to predict different event types with varied patterns.~We performed a t-test on $\alpha_1$, as shown in Table \ref{tab:case}. $\alpha_1$ refers to the weight computed in Equation~\eqref{eq7}, which is assigned to the historical expert for prediction. 
As shown in the `Mean' row, we observe that the mean value of $\alpha_1$ for historical events is relatively higher than that for non-historical events.~With standard deviations calculated from 7,371 samples, we conducted a t-test with the alternative hypothesis that `The mean weight for historical events is greater than that for non-historical events', which was validated with a highly significant p-value ($p < 0.001$). 

\begin{table}[t]
\centering
\begin{tabular}{cccc}
\hline
 $(M,N)$              & MRR   & H@3   & H@10    \\ \hline
(1,1)& 44.36&	49.81&64.21 \\
(1,2)& 43.94&	48.62&	63.03 \\
(2,1)      &44.08	&48.55	&63.00 \\ 
(2,2)      &43.91 & 48.60 & 62.81 \\ 
(3,1)  & 44.19 & 48.89 &62.87  \\
(1,3)      &43.91 & 48.79 & 63.13 \\ 
(3,3)  & 43.81 & 48.62 &62.94  \\
 \hline
\end{tabular}
\caption{Expert configuration tests on ICEWS14.}
\label{tab:expert_num}
\vspace{-0.3cm}
\end{table}
\vspace{+0.1cm}\noindent\textbf{Sensitivity Analysis on Event-aware Experts Configuration.}\label{app:sense}
In this part, we analyze the experiment results of varying the number of historical/non-historical expert modules. As shown in Table~\ref{tab:expert_num}, the optimal performance is achieved with $(M,N)=(1,1)$. As the number of experts increases, the prediction performance tends to decrease, indicating that complex combinations of expert modules are not necessary for the TKG Reasoning task. In fact, increasing the number of experts may lead to parameter redundancy and raise the risk of overfitting.

\section{Conclusion}
In this work, we proposed a Multi-Expert Structural-Semantic
Hybrid (MESH) framework.~Through the design of expert modules and the event-aware gate function, our model enabled adaptive information fusion based on event types.~Extensive experiments on three public datasets demonstrated that our approach consistently outperformed existing methods.
\section*{Limitations}
The performance of MESH relies on the effectiveness of the underlying structural encoder and semantic encoder. The quality of event representations and the final prediction results are bounded by the capacity of these encoders. In the future, the framework can incorporate more advanced feature encoders and enhance its performance. The modular design allows MESH to benefit from future improvements in representation learning techniques while maintaining its core architecture.
\section*{Acknowledgments}
This work is in part funded by the National Natural Science Foundation of China (No. 62372364); in part by Shaanxi Provincial Technical Innovation Guidance Plan, China under Grant 2024QCY-KXJ-199; in part by Research Impact Fund (No.R1015-23), Collaborative Research Fund (No.C1043-24GF) and Tencent (CCF-Tencent Open Fund, Tencent Rhino-Bird Focused Research Program).
\bibliography{acl}
\clearpage
\appendix
\newpage

\begin{algorithm}[t]
\caption{Training Process of MESH}
\label{train}
\begin{algorithmic}[1]
\REQUIRE Training set $\mathcal{D}$, expert weight $\omega$, pre-trained language model $LLM$ 
\ENSURE Optimized model parameters $\theta^*$

\STATE \textbf{Stage 0:} Pre-training
\STATE Train structural encoder $G$ on $\mathcal{D}$
\STATE Freeze parameters of $G$

\STATE \textbf{Stage 1:} Main Training Process
\STATE Load $G$, $LLM$ and obtain entity and relation representation $\boldsymbol{H}_g,\boldsymbol{R}_g,\boldsymbol{H}_{LLM},\boldsymbol{R}_{LLM}$ (Eq.~\ref{eq1}) 
\STATE Initialize model parameters $\theta$
\WHILE{Not Convergence}
    \FOR{event $(s,r,o,t) \in \mathcal{D}$}
    \STATE Obtain $\boldsymbol{h}_g,\boldsymbol{r}_g$ and $\boldsymbol{h}_l,\boldsymbol{r}_l$ (Eq.~\ref{eq2})
    \STATE Calculate structural and semantic query representation $\boldsymbol{q}_g,\boldsymbol{q}_l$ (Eq.~\ref{eq3},~\ref{eq4})
    \STATE Calculate historical identifier $I^{s,r}_t(o)$ (Eq.~\ref{eq15}), historical/non-historical query representation $\boldsymbol{q}^{his},\boldsymbol{q}^{nhis}$ (Eq.~\ref{eq10}, ~\ref{eq11})
    \STATE Calculate final query representation $\boldsymbol{q}$ (Eq.~\ref{eq:predict-dot})
        \STATE Make final prediction $\boldsymbol{p}_{s,r,t}$ and expert predictions $\boldsymbol{p}_{s,r,t}^{his}$, $\boldsymbol{p}_{s,r,t}^{nhis}$  (Eq.~\ref{eq:finalP},~\ref{eq12}, ~\ref{eq13})
        \STATE Compute expert losses $\mathcal{L}_e^{his}$, $\mathcal{L}_e^{nhis}$,
        prediction loss $\mathcal{L}^m$ (Eq.~\ref{eq16}, ~\ref{eq17}, ~\ref{eq18})
    \STATE  Get total loss $\mathcal{L}$ (Eq.~\ref{eq19})
        
        \STATE Update parameters via backpropagation
        \ENDFOR
    \ENDWHILE

\RETURN Optimized model parameters $\theta^*$
\end{algorithmic}
\end{algorithm}

\section{Optimization Algorithm}\label{app:Opt}
To better illustrate the optimization of MESH, we present the complete training process in Algorithm~\ref{train}, which shows the two-stage training process and the integration of expert modules. First, we train and freeze the parameters of the structural encoder $G$ (lines 2-3). Then, we obtain entity and relation representations from both $G$ and $LLM$ to derive the structural and semantic query representations (lines 5, 9-10).~Next, we employ event-aware experts to integrate semantic and structural information, followed by a prediction expert to combine the outputs from event-aware experts, generating integrated query representations and predictions (lines 10-13). Finally, we compute the expert loss to guide the model in learning different patterns for different event types and prediction loss to enhance its reasoning capability, then update the model parameters (lines 14-16).


\section{Experiment Setups}
\subsection{Evaluation Metrics}\label{metric}
In this section, we introduce the formal definitions of MRR and H@k metrics.
\begin{equation}
    \text{MRR} = \frac{1}{|Q|} \sum_{i=1}^{|Q|} \frac{1}{\text{rank}_i}
\end{equation}
where $|Q|$ denotes the total number of queries and $\text{rank}_i$ represents the rank position of the correct answer for the $i$-th query.

\begin{equation}
    \text{Hit@k} = \frac{1}{|Q|} \sum_{i=1}^{|Q|} \phi(\text{rank}_i \leq K)
\end{equation}
where $\phi(\cdot)$ is an indicator function that returns 1 if the condition is satisfied and 0 otherwise.

\begin{algorithm}[t]
\caption{Naive Approach}
\label{alg:historical}
\begin{algorithmic}[1]
\REQUIRE Test set $\mathcal{D}_{test}$, training set $\mathcal{D}_{train}$, rank threshold $k$
\ENSURE Hits@$k$ scores
\FOR{each $(s,r,o,t) \in \mathcal{D}_{test}$}
    \IF{$\exists (s,r,*,t') \in \mathcal{D}_{train}$}
        \STATE Count frequency of objects interacting with $(s,r)$
    \ELSE
        \STATE Count frequency of objects interacting with $s$
    \ENDIF
    \STATE $rank \gets$ Position of $o$ in frequency-sorted list
    \IF{$rank \leq k$}
        \STATE $hits@k \gets hits@k + 1$
    \ENDIF
\ENDFOR
\RETURN $hits@k/|\mathcal{D}_{test}|$
\end{algorithmic}
\end{algorithm}

\subsection{Lightweight Baselines}\label{naive}
 To investigate the performance of simple baselines in the TKG reasoning task, we propose two lightweight approaches based on dataset statistics or large language model encoding. 
 
 The approach ``Naive'' simply selects the top k entities that have the most interactions with the query or subject as the prediction results, as shown in Algorithm \ref{alg:historical}. This method is training-free and the execution time is less than 10 seconds.~As shown in Table~\ref{tab:main results transductive}, it achieves comparable performance to some graph-based methods on both ICEWS14 and ICEWS05-15.
 
 The approach LLM-MLP employs MLP layers to compress the embeddings of entities and relations obtained from LLMs, then utilizes a ConvTransE decoder for prediction, reducing inference costs by more than 90\% compared to existing LLM-based approaches. However, it performs well across three datasets compared to other generative LLM-based methods. Specifically, it outperforms ICL across most of metrics. 

 Experiments of these two approaches reveal that TKG reasoning often follows repetitive patterns. However, it is also important to capture the emergence of new events (non-historical events).

\section{Additional Experiments}

\subsection{Ablation on Gating Input}\label{app:ablation}
In this section, we conduct experiments of using different inputs~(structural, semantic, and concatenated) for the query-motivated gate to compare how different types of information guide the wight allocation and affect prediction results.~The experimental results in Table~\ref{tab:input} demonstrate that using structural information as input achieves the best performance, because structural information not only contains the information of the query context, but also includes global knowledge of the graph, which can better guide the gate's output. 
\begin{table}[t]
\centering

\begin{tabular}{cccc}
\hline
 Input              & MRR   & H@3   & H@10    \\ \hline
Structural& 44.36&	49.81&64.21 \\
Semantic      &44.12 & 49.31 & 63.26 \\ 
Concatenated  & 44.17 & 49.40 &63.57  \\
 \hline
\end{tabular}
\caption{Gate input analysis on ICEWS14.}
\label{tab:input}
\end{table}

\begin{table}[t]
\centering
\vspace{-0.2cm}
\begin{tabular}{cccc}
\hline
 Methods             & MRR   & H@3   & H@10    \\ \hline
RE-GCN	&19.16	&20.41&	33.10 \\
LLM-MLP	&15.71	&15.89&	26.33 \\
MESH(ours)	&19.96&	21.43&	34.37 \\ 

 \hline
\end{tabular}
\caption{Experiments on GDELT dataset.}
\label{tab:gdelt}
\vspace{-0.3cm}
\end{table}
\subsection{Experiments on GDELT Dataset}
We have conducted experiments on one more dataset, GDELT, from news domain, which has a finer temporal granularity (15 minutes). It is a global event dataset based on international news reports. 
We compare our proposed method with graph-based baseline RE-GCN and LLM-based baseline LLM-MLP on the GDELT~\cite{leetaru2013gdelt} dataset.~Experimental results in Table~\ref{tab:gdelt} show that our framework, by jointly leveraging structural and semantic information, consistently outperforms approaches that utilize only one type of information.

\begin{table}[t]
\centering
\vspace{-0.2cm}
\begin{tabular}{cccc}
\hline
       MESH      & MRR   & H@3   & H@10    \\ \hline
ICEWS14, 100\%	&44.36	&49.81	&64.21 \\
ICEWS14, 75\%	&43.91	&48.62	&63.62 \\
ICEWS14, 50\%	&42.24	&46.6	&61.02 \\ 

 \hline
\end{tabular}
\caption{Experiments on incomplete historical data.}
\label{tab:remove}
\vspace{-0.3cm}
\end{table}
\subsection{Experiments on Incomplete Historical Data}
To simulate scenarios with incomplete historical data, we randomly removed 25\% and 50\% of historical event records from the ICEWS14 training set and conducted experiments using our method MESH. The results in Table \ref{tab:remove} show that although reducing historical data slightly degrades reasoning performance, our approach remains robust, highlighting both the importance of historical events and our model’s resilience to missing information.

\end{document}